\documentclass[a4page,10pt]{article}
\usepackage[top=0.8in, bottom=0.8in, left=0.5in, right=0.5in]{geometry}
\usepackage{fancyhdr}
\renewcommand{\thispagestyle}[1]{}
\usepackage[pdftex]{graphicx}
\usepackage[utf8]{inputenc}
\usepackage{hyperref}
\usepackage{tabularx}
\usepackage{floatrow}
\newfloatcommand{capbtabbox}{table}[][\FBwidth]

\title{Deep Convolutional Neural Networks for Pairwise Causality}

\begin{document}
\author{Karamjit Singh, Garima Gupta, Lovekesh Vig, Gautam Shroff, and Puneet Agarwal
%
%
\vspace{.3cm}\\
%
TCS Research, New-Delhi, India\\
%
}
\maketitle 

\begin{abstract}
\label{Abs}
Discovering causal models from observational and interventional data is an important first step preceding what-if analysis or counterfactual reasoning. As has been shown before\cite{fonollosa2016conditional}, the direction of pairwise causal relations can, under certain conditions, be inferred from observational data via standard gradient-boosted classifiers (GBC) using carefully engineered statistical features. In this paper we apply deep convolutional neural networks (CNNs) to this problem by plotting attribute pairs as 2-D scatter plots that are fed to the CNN as images. We evaluate our approach on the `Cause-Effect Pairs' NIPS 2013 Data Challenge\cite{guyon2014results}. We observe that a weighted ensemble of CNN with the earlier GBC approach yields significant improvement. Further, we observe that when less training data is available, our approach performs better than the GBC based approach suggesting that CNN models pre-trained to determine the direction of pairwise causal direction could have wider applicability in causal discovery and enabling what-if / counterfactual analysis. 
\end{abstract}

\section{Introdution}
It is well known that both What-if analysis and Counterfactual reasoning can be performed using \textit{causal inference} on Causal Bayesian networks via Intervention and Counterfactual queries\cite{koller2009probabilistic}\cite{pearl2009causality}. Causal discovery is an important first step to learn Causal Bayesian networks.
The extent to which the direction of edges in a such a causal network can be identified from purely observational data is limited. Techniques
based on conditional independence tests can only discover edge directions within the limits of Markov equivalence. Statistical techniques for
determining the direction of a pairwise causal relationship, such as \cite{fonollosa2016conditional} can, in certain conditions, augment the causal discovery process.
In this paper, we use deep convolutional neural networks (CNNs)\cite{krizhevsky2012imagenet} to \textbf{improve the state of art} in detecting the direction of pairwise
causal relationships from purely observational data.
\begin{figure}[ht]
\centering
\includegraphics[width=60mm]{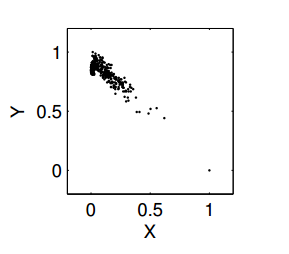}
\caption{Example of a pairwise causal discovery task: decide whether $X$ causes $Y$ , or $Y$
causes $X$, using only the observed data (visualized here as a scatter plot).}
\label{fig:scatter}
\end{figure}

We are concerned with determining the directions of a causal relationship between two attributes $X$ and $Y$, i.e., the problem is to identify whether $X$ causes $Y$ or, alternatively, $Y$ causes $X$, given joint observations of two attributes.
In general, the identification of causal relationships, including the pairwise case, requires controlled experimentation, which in many cases, is too expensive, unethical, or technically impossible to perform~\cite{mooij2010distinguishing}. The development of methods to identify causal relationships from purely observational data 
is therefore desirable, and also challenging: Consider the data visualized in Figure\ref{fig:scatter} in  the form of scatter plot. The question is: does X cause Y , or does Y cause X? The true answer is $X$ causes $Y$, as here X is the altitude of weather stations and Y is the mean temperature measured at these weather stations\cite{mooij2010distinguishing}. In the absence of knowledge about the measurement procedures that the variables correspond with, one can exploit statistical patterns in the data in order to find the causal direction. Approaches to causal discovery based on conditional independences do not work here, as $X$ and $Y$ are typically dependent, and there are no other observed variables to condition on.

Many causal discovery methods have been proposed in recent years that were claimed to solve this task under various assumptions\cite{fukumizu2007kernel}\cite{kano2003causal}\cite{stegle2010probabilistic}\cite{daniusis2012inferring}. Many of these approaches use the complexity of the marginal and conditional probability distributions in one way or another. Intuitively, the idea is that the factorization of the joint density $p_{C,E}(c, e)$ of cause C and effect E into $p_C(c)p_{E | C}(e | c)$ typically yields models of lower total complexity than the alternative factorization into $p_E(e)p_{C | E}(c | e)$. Alternative techniques use the distribution and independence properties of noise in the data, e.g. ANM\cite{hoyer2009nonlinear} that assumes that effects are non-linear functions of their causes plus independent non-gaussian noise, whereas, LINGAM\cite{shimizu2006linear} assumes functions are linear.

\textbf{Key Contributions:} 
In this paper we show that deep convolutional neural networks can discover pairwise causal relationships by merely looking at the scatter-plots
of data. Further, an ensemble of CNNs with the best available statistical techniques betters the state of the art in pairwise causal discovery
on the NIPS 2013 `cause-effect pairs' data challenge~\cite{guyon2014results} for this task. Finally, we show that when faced with sparse observations, our CNN-based
approach independently outperforms previous statistical approaches.\label{sec:intro}
%

\section{Approach}
We assume $n$ instances $D_1, D_2,..., D_n$ of data, where each $D_i$ contains $n_i$ pairwise observations ${(x^{(i)}_{1}, y^{(i)}_{1}), (x^{(i)}_{2}, y^{(i)}_{2}), ...,(x^{(i)}_{n_i}, y^{(i)}_{n_i})} $ for attributes $X_i$ and $Y_i$. These attributes can be continuous (numeric), binary, or categorical (cat). An attribute pair $(X_i, Y_i)$ is labeled as 1 if a causal relation exist from $X_i$ to $Y_i$, labeled as -1 if $Y_i$ to $X_i$, else labeled as 0. 
Our goal is to predict whether $X_i$ causes $Y_i$, or $Y_i$ causes $X_i$, or neither.

\begin{figure}[ht]
\centering
\includegraphics[width=120mm]{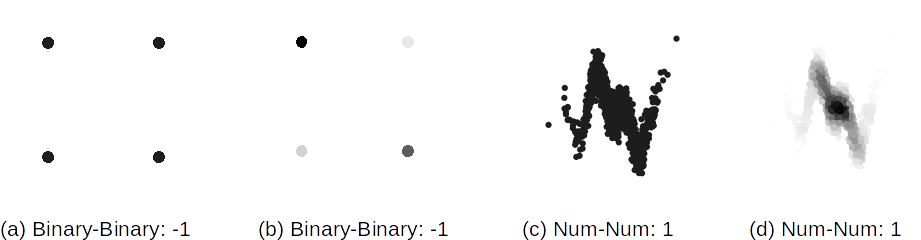}
\caption{Examples of four Scatter plots along with their labels. It shows the uses of normalized frequency in case binary or categorical attributes(Figure(a) and  (b)). Also, shows that for continous(num) attributes using normalized frequency may mask the important information(Figure(c) and (d))}
\label{fig:plots}
\end{figure}

For each $D_i$ we discretize values of $x_i$ and $y_i$ using $m$ bins:
Each data point in $D_i$ is mapped to a cell of a 2-D $m*m$ scatter plot.
For cases where neither $X_i$ nor $Y_i$ is continuous, i.e. when attributes are either binary or categorical, mapping an observation as a 2-D point on a scatter plot does not provide any information, e.g. As shown in Figure~\ref{fig:plots}(a), there are only four possible observations. For such cases, we calculate the normalized frequency of the occurrence of an observation and use it as a color intensity of a bin on a scatter plot (as shown in Figure~\ref{fig:plots}(b)), where highest frequency is mapped to 255 with the darkest colour and lowest frequency  is mapped to 0 with the white colour. For cases where either of the attribute is continuous, we map each observation as a 2-D point on a scatter plot(as  shown in Figure~\ref{fig:plots}(c)). We do not use frequency as a colour intensity in case of continuous attributes as this can often mask important information(as shown in Figure~\ref{fig:plots}(d)).

\subsection{CNN based approach}
\begin{figure}[ht]
\centering
\includegraphics[width=140mm]{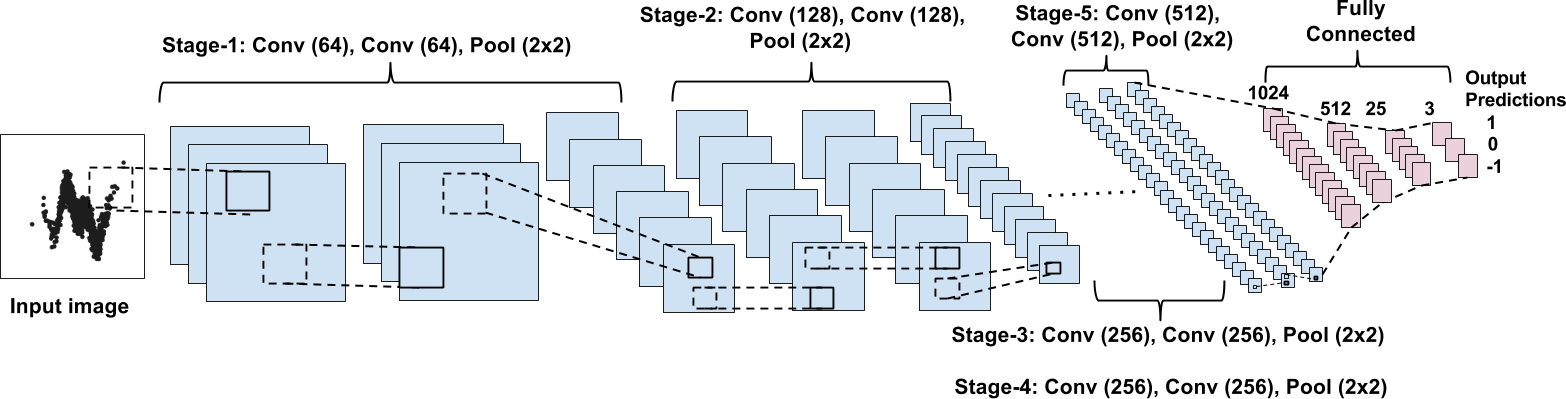}
\caption{Deep CNN Architecture used for our approach: 15 layers of convolution and pooling followed by three fully connected layers. It also shows the number of filters used at each layer.}
\label{fig:cnn}
\end{figure}

For each $D_i$, we generate a scatter plot of size 200*200 to train a CNN with architecture as in Figure~\ref{fig:cnn}: consist of 5 stages and at each stage we employ two convolutions and one pooling on the image which results into 15 layers of convolution and pooling followed by 3 fully connected layers with number of units as 1024, 512, and 25 respectively, and 3x3 filters. The output layer of CNN is a softmax layer with three units corresponding to the three classes with which we obtain three probabilities $p^c_1$, $p^c_0$, and $p^c_{-1}$ for each of the three classes.

\label{sec:causal}

\section{Experiments}
We consider the NIPS 2013 `cause-effect pairs' data challenge for the evaluation of our approach. The data consists of three parts: SUP1, SUP2, and CETrain. Type of attributes in each dataset and thier statistics are shown in Table~\ref{table:stats}. 
We randomly split the data into Training, Validation, and Test sets with ratio of 70:15:15 respectively to evaluate the vanilla CNN model as described above, as well as the following GBC based approach:

\renewcommand{\arraystretch}{1}
\begin{table*}[!b]
\caption{Shows number of instances in each dataset and the range of number of observation for all three datasets}
\label{table:stats}
\centering
\begin{tabular}{|c|c|c|c|}
\hline
Dataset & Type of attributes & Number of instances  & Number of observations \\
\hline
SUP-1 & continous & 5998 & 500 to 7988\\
\hline
SUP-2 & mixed & 5989 & 500 to 7998\\
\hline
CETrain & mixed & 4050 & 53 to 7998\\
\hline
Total & mixed & 16037 & 53 to 7998\\
\hline
\end{tabular}
\end{table*}

\textbf{GBC:} We use the code \footnote{https://github.com/jarfo/cause-effect} provided by the second winner of the data challenge. In their approach, they derive 43 features which include standard statistical features plus new measures based on variability measures of the conditional distributions and use these features to train GBC. We use the same code to generate features and train GBC on our training set.
See Table~\ref{table:gbc} for details. 
For each $D_i$ in test set, using the trained GBC, we obtain three probabilities $p^g_1$, $p^g_0$, and $p^g_{-1}$ for three classes.

\renewcommand{\arraystretch}{1}
\begin{table*}[!b]
\caption{Shows the list of hyper-parameters and their corresponding ranges used for tuning GBC. The best parameters obtained are shown in the rightmost column which are same as used in original code}
\label{table:gbc}
\centering
\begin{tabular}{|c|c|c|}\hline
& Grid search & Parameter selected\\
\hline
Number of estimators & 200,500,1000,1500 & 500\\
\hline
Maximum depth & 5,7,9,11,13 & 9\\
\hline
Minimum samples split & 8,20,100,200,400 & 8\\
\hline
Maximum features & `sqrt', `None' & None\\
\hline 
\end{tabular}
\end{table*}

\textbf{Weighted Ensemble:} For each $D_i$, we take the weighted sum of the probabilities obtained using CNN and GBC based approaches to generate three new probabilities as:
 $p^e_{k} = w*p^c_{k} + (1-w)*p^g_{k}$, where k = 0, 1, and -1. We validate the weight $w$ from the range $[0,1]$ with step size of 0.1 on validation set and use that weight for prediction on test set. The value of $w$ is $0.4$ in our case. 

Note that the symmetry of the task allows us to duplicate the training instances. Exchanging $X_i$ with $Y_i$ in an instance of label $c$ provides a new instance of the label $-c$. This idea has been used while learning both CNN and GBC mentioned above.

For a $D_i$, given the three probabilities $p_1, p_0,$ and $p_{-1}$ for three classes, we evaluate our approach mentioned above using two metrics:
(i) \textit{Accuracy:} We choose the class with the maximum probability as the predicted class for $D_i$ and we define \textit{accuracy} of a test set as the ratio of number of correctly predicted data instances to the total number of data instances in the test set.
and (ii) \textit{AUC:} We use the same metric as used in the classification task of the cause-effect pair data challenge\cite{cause}. 

\renewcommand{\arraystretch}{1}
\begin{table*}[!b]
\caption{Table showing Accuracy and AUC using CNN, GBC, and Ensemble for Validation, test and Kaggle test set}
\label{table:categories}
\centering
\begin{tabular}{|c|c|c|c|}
\hline
& GBC & CNN & Ensemble \\
\hline
Accuracy on Validation split & 77.8\% & 72.5\% & 79.3\%\\
\hline
AUC on Validation split & 0.805 & 0.779 & 0.831\\
\hline
Accuracy on Test split& 77.5\% & 73.3\% & 79.2\%\\
\hline
AUC on Test split& 0.81 & 0.769 & 0.833\\
\hline
\textbf{AUC on Kaggle Test data} & 0.81 & 0.7331 & \textbf{0.825} \\
\hline
\end{tabular}
\end{table*}

We evaluate our approaches for following two cases: In Case-1, we use full data given, i.e., all data instances and all observations of each data instance; and evaluate our CNN based and weighted ensemble approach on the test set and compare it with GBC based approach (evaluated on the same test set). In Case-2, we compare CNN with GBC based approaches for sparse training data by restricting number of observations in each instance.

\begin{figure}
\centering
\includegraphics[width=110mm]{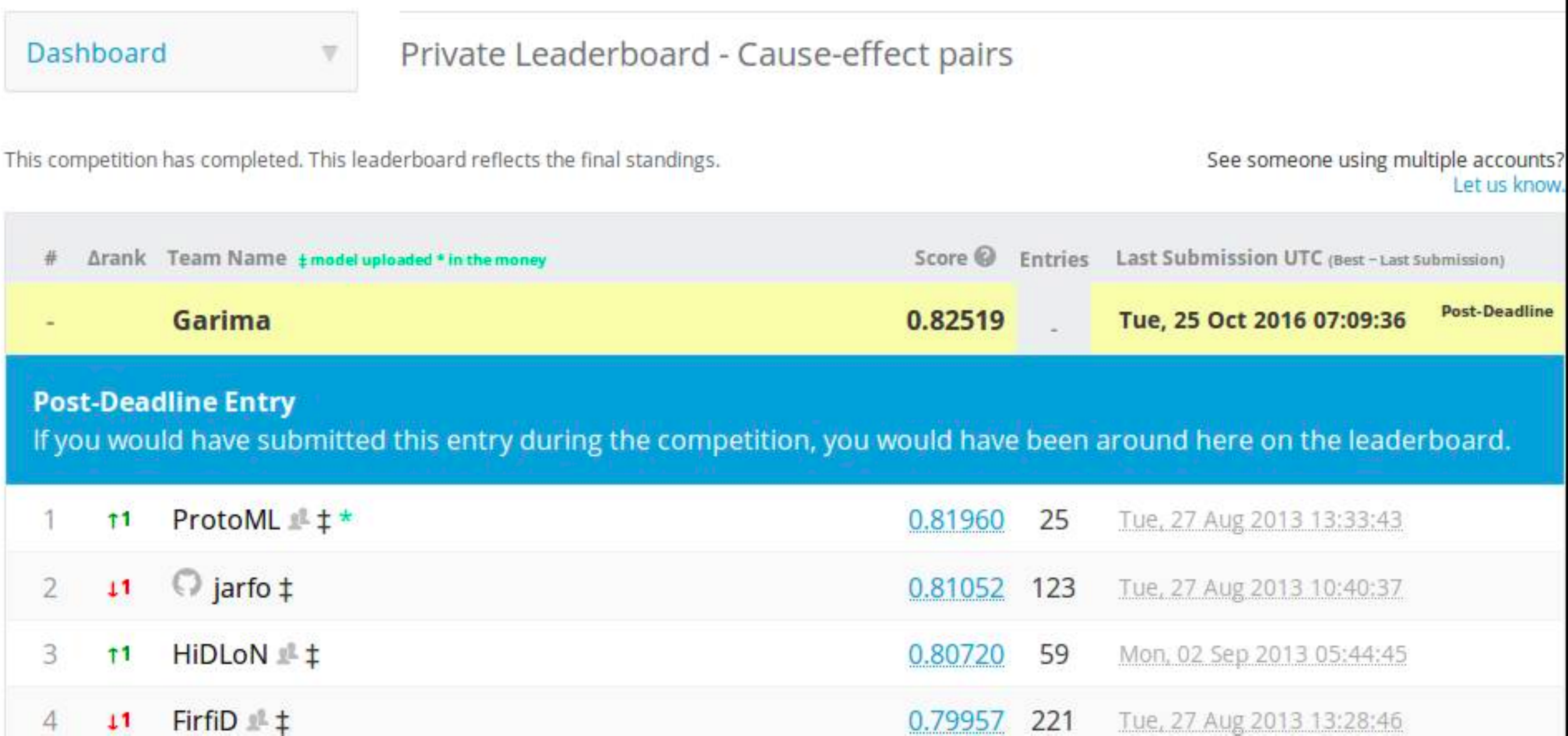}
\caption{AUC computed using our approach on test set available on Kaggle}
\label{fig:kaggle}
\end{figure}

\textbf{Case-1: Full Training Data:} Table~\ref{table:categories}, shows accuracy and AUC evaluated on Validation and Test set using CNN, GBC, and weighted Ensemble based approach. It shows that accuracy and AUC using CNN based approach is fairly good but not at par with the GBC based approach. From Table~\ref{table:categories}, it can be seen that the weighted ensemble of CNN and GBC gives a boost to both accuracy and AUC, suggesting that the two approaches of GBC and CNN are complementary to each other. Also, Table~\ref{table:categories} shows the AUC predicted using the three approaches on the test set given on the webpage of data challenge on the Kaggle website\footnote{https://www.kaggle.com/c/cause-effect-pairs/data}. An AUC of 0.825 is obtained using weighted ensemble based approach which \textbf{surpasses the state-of-the art}, getting us the first rank on the private leader-board (as shown in Figure~\ref{fig:kaggle}).

\begin{figure}
\centering
\includegraphics[width=90mm]{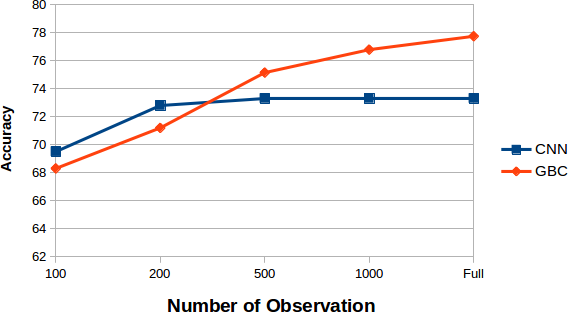}
\caption{Accuracy comparison of CNN and GBC for different number of observations}
\label{fig:acc}
\end{figure}

\begin{figure}
\centering
\includegraphics[width=90mm]{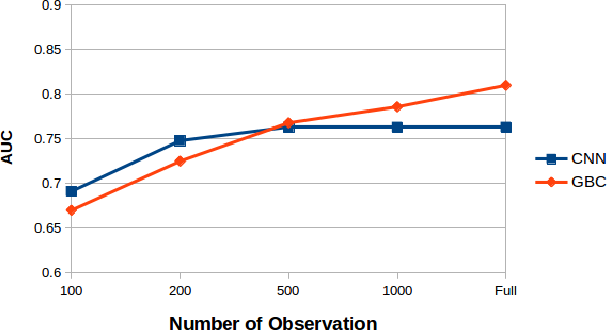}
\caption{AUC comparison of CNN and GBC for different number of observations}
\label{fig:auc}
\end{figure}

\textbf{Case-2: Sparse Training Data:} As shown in Table~\ref{table:stats}, number of observations in total data varies from 53 to 7998. In Case-2, we compare CNN with GBC based approach for the case where the amount of training data used for a particular attribute-pair is drastically reduced, a scenario that is often faced in real-life. We compare the accuracy and AUC using CNN and GBC for 100, 200, 500, and 1000 observations. As shown in Figure~\ref{fig:acc} and~\ref{fig:auc}, we observe that the CNN based approach outperforms GBC based approach when instances have few observations. 

Further, since CNN merely take scatter plots as input with no assumption about the data and no data specfic features, a model trained on one database can potentially be used for predicting causality on completely new dataset.

\label{sec:exp}
\section{Conclusion}
We propose the novel idea of using CNN to predict pairwise causality by merely looking at the scatter plots of data and show that the CNN based approach is complimentary to the existing statistical approach by improving the state of art through weighted ensemble. We also show the CNN based approach outperforms existing approach while given data has few observations.\label{sec:conclusion}

\bibliographystyle{unsrt}
\bibliography{esann}
\end{document}